\documentclass[final]{cvpr}

\usepackage{times}
\usepackage{epsfig}
\usepackage{graphicx}
\usepackage{amsmath}
\usepackage{amsmath,bm}
\usepackage{amssymb}

\usepackage{comment}
\usepackage{color}
\usepackage{booktabs}
\usepackage{multirow}
\usepackage{float}
\usepackage{wrapfig}
\usepackage{subfig}
\usepackage{marvosym}
\usepackage[table]{xcolor}
\usepackage[linesnumbered,ruled,vlined]{algorithm2e}
\usepackage{makecell}

\usepackage[pagebackref=true,breaklinks=true,colorlinks,bookmarks=false]{hyperref}

\def\cvprPaperID{3072} 
\def\confYear{CVPR 2021}

\begin{document}
\interfootnotelinepenalty=10000

\title{MagDR: Mask-guided Detection and Reconstruction for Defending Deepfakes}

\author{Zhikai Chen\\
Xi’an Jiaotong University\\
{\tt\small zhikai\_chen@outlook.com}
\and
Lingxi Xie\\
Huawei Inc.\\
{\tt\small 198808xc@gmail.com}
\and
Shanmin Pang\textsuperscript{\Letter} \\
Xi’an Jiaotong University\\
{\tt\small pangsm@xjtu.edu.cn}
\and
Yong He\\
Xi’an Jiaotong University\\
{\tt\small hy0275@stu.xjtu.edu.cn}
\and
Bo Zhang\\
Tencent Blade Team\\
{\tt\small cradminzhang@tencent.com}
}

\makeatletter
\let\@oldmaketitle\@maketitle
\renewcommand{\@maketitle}{\@oldmaketitle
  \begin{center}
  \includegraphics[width=0.91\linewidth]{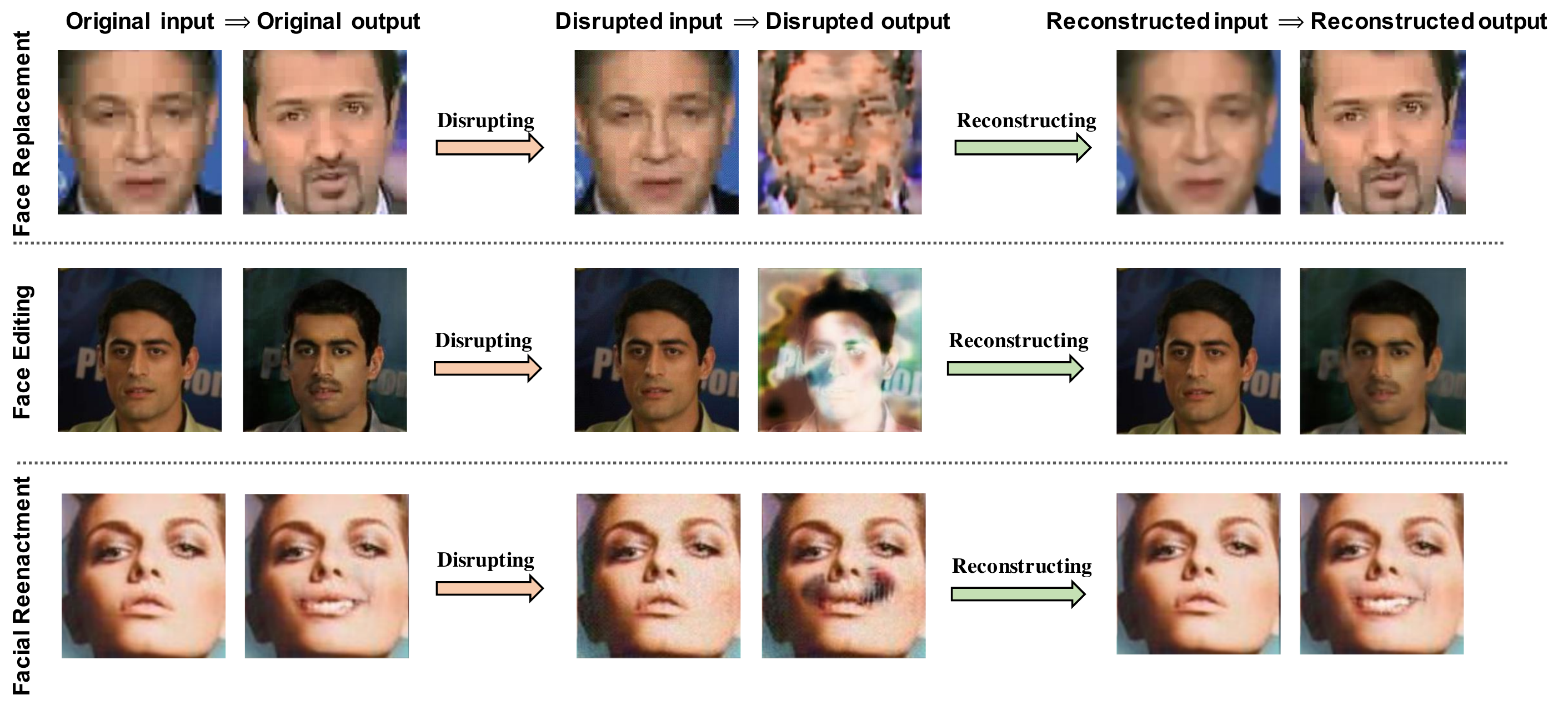}\\
  \end{center}
    \refstepcounter{figure}Figure~\thefigure: 
    MagDR defends deepfakes from adversarial attacks in three main tasks, face replacement (\textbf{top}), face editing (\textbf{middle}), and facial reenactment (\textbf{bottom}). In each group, we show the original generation effect (\textbf{left}), how adversarial perturbations damage the generation (\textbf{middle}), and how the proposed defender recovers the desired output (\textbf{right}).
    \label{fig:attack_performance}
    \bigskip\bigskip
    }
\makeatother\maketitle
\pagestyle{empty}
\thispagestyle{empty}
\begin{abstract}
\footnote{This work was supported by NSFC under Grant 61972312 and by the Key Research and Development Program of Shaanxi under Grant 2020GY-002.}Deepfakes raised serious concerns on the authenticity of visual contents. Prior works revealed the possibility to disrupt deepfakes by adding adversarial perturbations to the source data, but we argue that the threat has not been eliminated yet. This paper presents MagDR, a mask-guided detection and reconstruction pipeline for defending deepfakes from adversarial attacks. MagDR starts with a detection module that defines a few criteria to judge the abnormality of the output of deepfakes, and then uses it to guide a learnable reconstruction procedure. Adaptive masks are extracted to capture the change in local facial regions. In experiments, MagDR defends three main tasks of deepfakes, and the learned reconstruction pipeline transfers across input data, showing promising performance in defending both black-box and white-box attacks.  
\end{abstract}

\section{Introduction}
\label{sec:Intro}

Deepfakes originally appeared as a neutral technology that can synthesize images with the human face replaced by another identity. While the technique benefits the community in the scenarios of \textit{e.g.} creating new characters or decorate them with vivid facial expressions, it gradually becomes infamous for the unethical applications (\textit{e.g.} swap fake of celebrities into pornographic videos, or generate a fraud video that delivers fake and malicious messages). To avoid negative impacts to the public, researchers started to develop algorithms to detect the images and videos that have been contaminated by deepfakes~\cite{Rathgeb_2020_PRNU,zhang2019detecting}. However, the follow-up research~\cite{carlini2020evading,g2020adversarial,neekhara2020adversarial} quickly realized that these detectors are easily fooled by adversarial perturbations. Another way to confront deepfakes is to add adversarial perturbations to the source image so that the output is severely damaged~\cite{ruiz2020disrupting,Yeh_2020_WACV}. This was believed to be more robust than the deepfakes detectors.

However, in this paper, we reveal the feasibility of defending the adversarial attacks to deepfakes. We propose a framework named \textbf{mask-guided detection and reconstruction} (MagDR). It starts with defining a few criteria (\textit{e.g.}, SSIM, PSNR, \textit{etc.}) that are sensitive to the abnormality of the outputs. Then, a mask-guided detector is trained to judge, from the output image, whether the input image has been contaminated. If yes, a reconstruction algorithm follows to eliminate the damage of the adversarial perturbations and recover the desired output.

A highlight of our approach is that we maintain a number of masks and use them to provide auxiliary information in the detection and reconstruction procedures. The masks can be learned from an individual training process, and each of them corresponds to a specific part of the human face. Guided by the masks, the detector can be partitioned into two components which detect distortion and inconsistency, both of which indicate the regions that are likely to be contaminated. To reconstruct the desired output, we design a pipeline containing several modules and equip them with a changeable execution order and adjustable parameters. Then, we perform an adaptive optimization that suppresses all the pre-defined criteria and produce the recovered output.

We evaluate our approach on two popular datasets, namely, FaceForensics++~\cite{rossler2019faceforensics++} and CelebA~\cite{liu2015faceattributes}. We correspond three image-to-image translation methods, CycleGAN~\cite{zhu2017unpaired}, StarGAN~\cite{choi2018stargan}, and GANimation~\cite{pumarola2018ganimation}, to the three main functions of deepfakes, face editing, facial reenactment, and face replacement, respectively. We investigate two settings, one is the oblivious attack in which the attackers transfer the perturbations computed on original deepfakes models to the defender, and the other is the adaptive attack in which the parameters of the defenders are known to the attacker. Experiments show that deepfakes are vulnerable in both scenarios, but MagDR is able to eliminate the impact of the attack in most cases. Typical examples are shown in Figure~\ref{fig:attack_performance}. MagDR also shows advantages in extensive experiments against state-of-the-art adversarial attackers~\cite{carlini2017towards,aleks2017deep} and defenders~\cite{lu2017safetynet,metzen2017detecting,meng2017magnet,ruiz2020disrupting,samangouei2018defensegan,Mustafa_2020,dziugaite2016study,xie2017mitigating}. Interestingly, MagDR is able to transfer across different scenarios, demonstrating its ability in both black-box and white-box attacks, and implying that adversarial perturbations are detectable by some common rules.

The contributions of this paper are as follows:
\begin{itemize}
\vspace{-0.2cm}
\setlength{\itemsep}{-0.2ex} 
\item We reveal that the threats of deepfakes have not yet been eliminated by adding adversarial perturbations to the input image or videos.
\item We find that the corruption to image-to-image translation can influence either a part of the image or the entire image. The proposed mask-guided design follows this property and achieves satisfying performance.
\item We propose a heuristic, hierarchical reconstruction module for each conditional attribute patch. We adjust it through a progressive approach, which can largely reduce the computational costs. Therefore, we verify that different regions are complementary in recovering the detailed textures, and the layer-by-layer architecture with a proper execution order can enhance the performance of defense.
\end{itemize}

\section{Related Work}
\textbf{Deepfakes.}
Deepfakes have gained a lot of concern for it can generate fake images, video, voice, \textit{etc}. Those generated products can highly mislead the judgment of humans. \cite{mirsky2020creation, tolosana2020deepfakes} survey deepfakes, which divide deepfakes on facial image or video into four main regions: Face Synthesis, Face Editing, Facial Reenactment, and Face Replacement. 
Face synthesis can generate entire non-existent face images~\cite{Karras_2019_CVPR,2019_Arxiv_GANRemoval_Tolosana,Jain2019facialManipulation} that usually use GAN based methods \textit{e.g.}, ProGAN~\cite{pgan} or StyleGAN~\cite{Karras_2019_CVPR}, \textit{etc}. 
Face Editing means some attributes of the face can be added, removed, or changed. Those attributes can be the hair, age, clothes, ethnicity, gender~\cite{2018_TIFS_SoftWildAnno_Sosa}, \textit{etc}. And the methods often related with GAN with attributes \textit{e.g.}, StarGan~\cite{choi2018stargan}, attGAN~\cite{he2019attgan} and STGAN~\cite{Liu_2019_CVPR}. 
Facial Reenactment modifying the facial expression of the person can be achieved by ~\cite{Liu_2019_CVPR,thies2016face2face,thies2019deferred}. 
Finally, face replacement~\cite{nirkin2019fsgan,li2019faceshifter} is an operation to swap the face of the source image to the target image by considering the face size, pose, and skin color \textit{etc}.
In this paper, we mainly focus on defending face editing, facial reenactment, and face replacement. This is because that these deepfakes alter source images, while face synthesis does not need any input images.

\textbf{Adversarial Attack and Defense.}
Researchers designed a lot of attacking algorithms to add imperceptible perturbations onto well-trained neural networks so that the prediction is dramatically destroyed.
Successful scenarios include image classification~\cite{kurakin2016adversarial,Moosavi_Dezfooli_2016,aleks2017deep}, object detection and semantic segmentation~\cite{Xie_2017}, image captioning~\cite{xu2019exact}, video classification~\cite{chen2019appending} \textit{etc}.
Among the first to introduce adversarial examples against deep neural networks was \cite{szegedy2013intriguing}. 
After that, Goodfellow \textit{et al.}~\cite{goodfellow2014explaining} used the sign of the gradient to propose a fast attack method called Fast Gradient Sign Method (FGSM). 
FGSM seeks the direction that can maximize the classification errors to update each pixel. 
In \cite{aleks2017deep}, an iterative method called Projection Gradient Descent (PGD) was proposed.  
PGD makes the perturbations project back to the $\epsilon$-ball which center is the original data when perturbations over the $\epsilon$-ball.
There have been a lot of adversarial defense strategies~\cite{advsurvey} (Adversarial Detecting\cite{meng2017magnet,lu2017safetynet,metzen2017detecting}, Input Reconstruction\cite{meng2017magnet,Xu_2018,Liang_2019}, Adversarial (Re)training\cite{kurakin2016adversarial,tramer2017ensemble}, \textit{etc}). 

\textbf{Attacking and Defending Deepfakes.} A lot of deepfake detectors are proposed to detect fake images in Face Synthesis \cite{stehouwer2019detection}, Face Editing \cite{Rathgeb_2020_PRNU,zhang2019detecting}, Facial Reenactment\cite{wang2020cnn} and Face Replacement \cite{li2020face}.
Despite popularity, recent researches \cite{carlini2020evading,g2020adversarial,neekhara2020adversarial} found those deepfake detectors are easily to be misled via adversarial perturbations. 
Another method to confront deepfakes was proposed by \cite{ruiz2020disrupting, Yeh_2020_WACV}. They found that similar to other traditional computer vision systems, deepfakes are also vulnerable to adversarial examples. 
Through adding adversarial perturbations to source images, the output can be corrupted, highly influencing the effectiveness of deepfakes.
While in this paper, we demonstrate that there is a method to resolve the newly proposed disrupting methods.

\def\loss{\bm{\mathcal{L}}}

\begin{figure*}[!tb]
\centering\includegraphics[width=\textwidth]{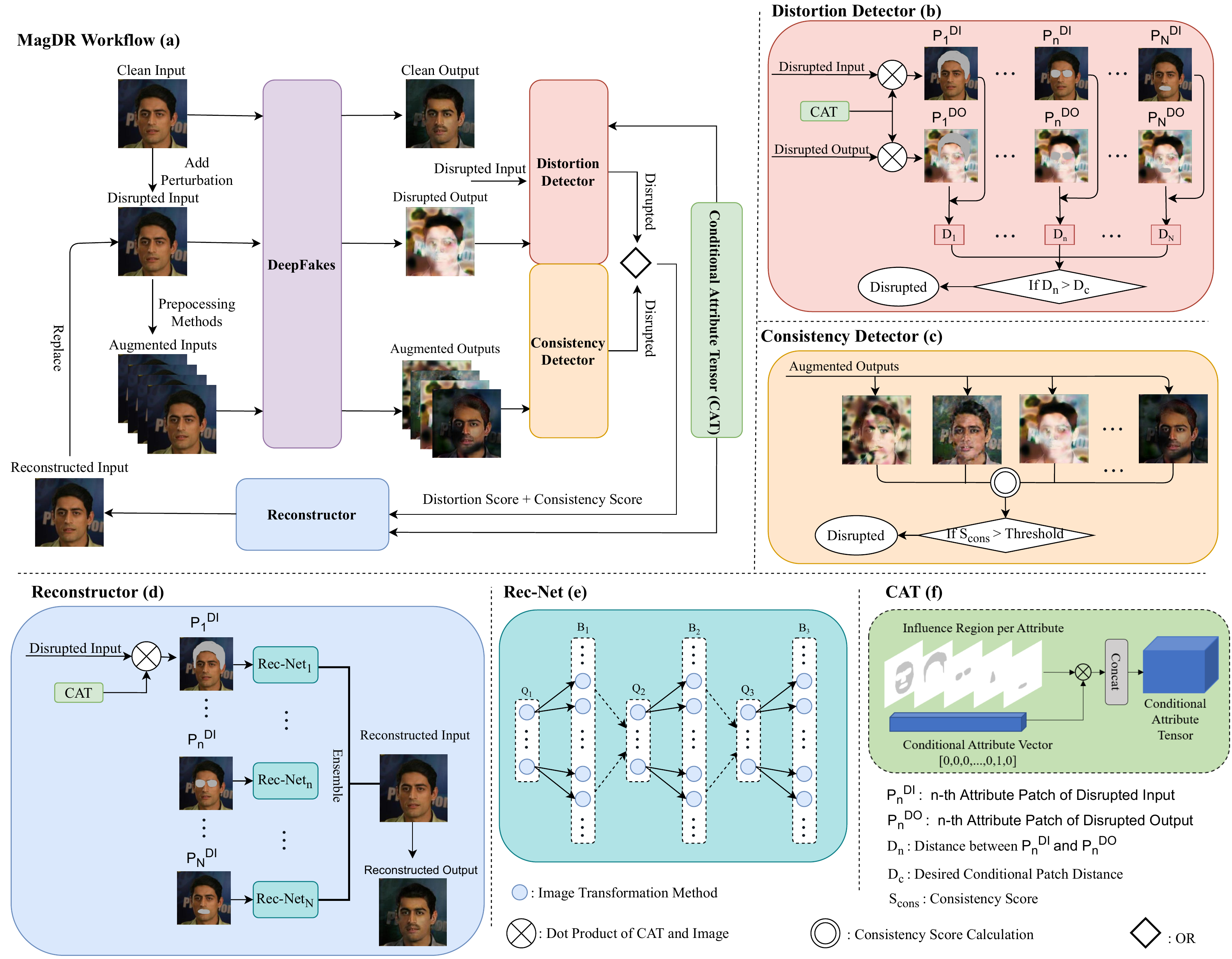}
\caption{\textbf{ The MagDR Framework.} It is a unified framework suitable for various deepfake models, \textit{e.g.}, StarGAN, and GANimation. Input with adversarial perturbations is first fed into detectors. If it is considered disrupted, MagDR reconstructs it through reconstructor, replace the disrupted input with reconstructed one, then do the processes again. The detector consists of two sub-detectors, and the core of the reconstructor is Rec-Nets. We use the pre-trained conditional attribute tensor to help detector and reconstructor for detailed image information, where the grey region of masks needs to be preserved when doing calculations.}
\label{fig:MagDR}
\vspace{-4mm}
\end{figure*}

\section{Methodology}

\subsection{An Overview of Deepfakes and Disrupted Images Generation}

CycleGAN~\cite{zhu2017unpaired} uses two sets of GANs, in which two Generators transform the images from both domains, \textit{i.e.}, $\mathbf{G_x : x \rightarrow y}$ and $\mathbf{G_y : y \rightarrow x}$, and two Discriminator $\mathbf{D_x}$ and $\mathbf{D_y}$ learn to distinguish between $\mathbf{x}$ and $\mathbf{G_Y(y)}$ as well as between $\mathbf{y}$ and $\mathbf{G_X(x)}$. An enhanced approach named StarGAN~\cite{choi2018stargan}, which performs image-to-image translations for multiple domains using only a single model. It is a conditional attribute transfer network trained by attribute classification loss and cycle consistency loss. Another work called GANimation~\cite{pumarola2018ganimation} reenacts face and uses an expression prediction loss to penalizes $\mathbf{G}$ for realistic expressions. In this model, the output of deepfakes can be simplified as $\mathbf{G(x, c)}$, where $\mathbf{c}$ is the target class that defines the specific condition where we want to modify in different deepfakes.

Generating disrupted images, which adds some imperceptible perturbation in the source image, is similar to generating adversarial examples. Recently, \cite{ruiz2020disrupting, Yeh_2020_WACV} utilized iterative gradient-based methods  (\textit{e.g.} I-FGSM~\cite{kurakin2016adversarial}) to generate disrupting images. In this work, we use both the iterative gradient-based method PGD~\cite{aleks2017deep} and the optimize-based strategy C\&W~\cite{carlini2017towards} to generate disrupted images for comprehensive attack settings.

Let $\mathbf{\hat{x}}$ be a generated disrupted input image, \textit{i.e.}, $\mathbf{\hat{x}} = \mathbf{x} + \mathbf{\delta}$, where $\mathbf{x}$ is the input image and $\mathbf{\delta}$ is a human-imperceptible perturbation. 
As such, the disrupted output can be formulated as $\mathbf{G(\hat{x}, c)}$, and the objective function for perturbation generation is:

\begin{equation}\label{eq:1}
    \max_{\mathbf{\delta}}\loss{}(\mathbf{G(\hat{x}, c)}, \mathbf{r}) \text{, ~~~}  
    \text{subject to } ||\mathbf{\delta}||_{\infty} \leq \epsilon,
\end{equation}
where $\mathbf{\epsilon}$ is the maximum magnitude of the perturbation and $\loss{}$ is the loss function to define the difference between the inputs. If we pick $\mathbf{r}$ to be the ground-truth output, \textit{i.e.}, $\mathbf{r = G(x, c)}$, we will get the \textit{ideal} disruption which aims to maximize the distortion of the output.

While sometimes we may need to get some specific altered output image. Accordingly, we need to minimize the distance $\loss{}$ between $\mathbf{G(\hat{x}, c)}$ and $\mathbf{r_t}$, where $\mathbf{r_t}$ can represent any image we want it to be. This is known as the targeted attack and its formulation can be organized as:

\begin{equation}\label{eq:2}
    \min_{\mathbf{\delta}}\loss{}(\mathbf{G(\hat{x}, c)}, \mathbf{r_t}) \text{, ~~~}  
    \text{subject to } ||\mathbf{\delta}||_{\infty} \leq \epsilon,
\end{equation}

In addition, to disrupt a network in many defense scenarios, we perform a modified $\loss{}$ calculation method that corresponds to \emph{adaptive attacks}~\cite{carlini2017adversarial,carlini2019evaluating,herley2017sok}:

\begin{equation}\label{eq:3}
    \max_{\mathbf{\delta}}\sum_{k=1}^K\loss{}(\mathbf{f}_k(\mathbf{G(x + \delta, c), r))} \text{, ~~~}  
    \text{subject to } ||\mathbf{\delta}||_{\infty} \leq \epsilon,
\end{equation}
where $\mathbf{f}_k$ is a defense pre-processing operation, and we have $K$ different defense methods with different magnitudes and types. 

\subsection{The Framework for Defending Deepfakes}

The proposed framework is named MagDR, standing for mask-guided detection and reconstruction. As shown in Figure~\ref{fig:MagDR} (a), it contains two major components, a detector and a reconstructor, both of which are guided by a set of pre-defined criteria computed on adaptive masks. The overall idea is to sense the presence of adversarial attacks from the output image (which is often significantly perturbed), and perform an adjustable algorithm to suppress all the criteria to an acceptable value, after which we believe that the output has been reconstructed.

Before entering the elaboration of technical details, we point out that the implementation of each module can be freely changed under the designed framework. That being said, we look forward to future research that improves the performance of MagDR by using more accurate criteria as well as stronger detectors and reconstructors.

\subsubsection{Predefined Module for MagDR}
\label{sec:pre}

\textbf{Distance Metrics Definition.} We define $\mathbf{D}$ as the whole set of distance estimation functions. Each $\mathbf{D}_i(\mathbf{x}_i,\mathbf{x}_j)$  represents a distance estimation function which calculates the  difference of input image pair. 
In particular,  $\mathbf{D}_i$ can be computed in a number of different ways. 
In our settings, we compute the distances between $\mathbf{\Phi}_l(\mathbf{G(\hat{x},c),r))}$ ~and ~$\mathbf{\Phi}_l(\mathbf{G(x,c),r))}$ on the $l$-th layer  using  $L_p$, SSIM and PSNR, where $\mathbf{\Phi}_l$ is the mapping from an image to its internal DNN representation at layer $l$.
Besides, we  calculate the cosine similarity at layer $L-1$, where $L$ denotes the number of layers of the network.

\textbf{Conditional Mask Tensor Generation.} 
For enhancing the detection and reconstruction, we bring 19-class soft facial region masks in our MagDR from a pre-trained face parser same as MagGAN~\cite{Wei_2020_ACCV} introduced. The face parser is a modified BiseNet~\cite{yu2018bisenet} trained on the CelebAMask-HQ dataset~\cite{lee2019maskgan}~\footnote{\url{https://github.com/zllrunning/face-parsing.PyTorch}}.
Then we select $N$ attribute region masks based on our deepfake tasks.
For each attribute $a_i$, we define its {\em influence regions} represented by two probability masks $M_i \in [0,1]^{H\times W}$. 
Then, we concatenate these attribute mask regions into a conditional mask tensor, where  the mask can be denoted as a probability map $\mathbf{M} \in [0,1]^{N \times H \times W}$ of the $N$ facial parts, satisfying $\sum^{N}_{i=1} \mathbf{M}_{i,h,w} = \mathbf{1}_{h,w}$. 
The module of  conditional mask tensor generation is shown in Figure~\ref{fig:MagDR} (f).

\subsubsection{The Detector}

As Figure~\ref{fig:MagDR} (a) shown, to detect different patterns of corruption, our detector contains two sub-detectors: a distortion detector and a consistency detector. We combine these two sub-detectors together as a function $d: \mathbb{S} \rightarrow \{0, 1\}$ to decide whether the input is an adversarial sample or not. 

\textbf{Distortion Detector.}  
As Figure~\ref{fig:attack_performance} shows, deepfakes normally  modify the conditional attribute region of the input. Thus, it is weird for the output that there is abnormality out of the conditional attribute region.
Motivated by this observation, we use the changes of attribute regions measured by the aforementioned conditional attribute region mask and distance metrics  to develop a distortion detector, as shown in Figure~\ref{fig:MagDR} (b).

\begin{equation}\label{eq:4}
    \bar{\mathbf{d}_i} = \frac{\mathbf{D}(\mathbf{M}_i\circ x, \mathbf{M}_i\circ \mathbf{G(x, c))}}{S_i}
\end{equation}
where $\bar{\mathbf{d}_i}$ is the distance metrics vector with the conditional attribute region, and $S_i$ is the number of pixels of the attribute region mask.

In practice, we cannot determine whether the image is polluted or not by directly using the distance of the target attribute region larger than a threshold. 
This is because that the threshold is affiliated with specific deepfake tasks and images. 
To address the problem, we propose to use the distance metrics of the target attribute region as the benchmark to compare with other regions. 

\begin{equation}\label{eq:5}
\mathbf{V}_{i} = 
\begin{cases}
1 & \textrm{if } \max \{\bar{\mathbf{d}_i} - \mathbf{d}_c\} \geq 0, \\
0 & \textrm{otherwise}
\end{cases}
\end{equation}
where $\mathbf{d}_c$ is the distance of the target attribute region $M_c$. 
Based on the difference of conditional regions, we can conclude the input is disrupted or not. 
In particular, if the distance metrics of the calculated region is larger than the benchmark, we will assume this region is corrupted and set the flag as 1. Otherwise, it is considered as clean and set as 0. Thus,  we naturally use the number of disrupted patches to decide whether an image is disrupted or not. 

In addition, we can use the vector $\bar{\mathbf{d}_i}$ to define distortion score that measures the distortion magnitude of the disrupted output. Specifically, it is  formulated as:

\begin{equation}\label{eq:8}
    \mathbf{S}_\mathrm{dist} = \sum_{i\ne c} \mathbf{w} \circ \text{sigmoid} (\mathbf{\bar{d}}_i)
\end{equation}
where  $\mathbf{w}$ is the regularization vector of the distance metric $\bar{\mathbf{d}_i}$.

\textbf{Consistency Detector.} The distortion detector is effective in detecting disrupted images when there are a lot of corruptions out of the conditional attribute region.
However, it becomes less effective if corruption is caused in the whole image. 
To overcome this problem, we consider the vulnerability of adversarial perturbations as a breakthrough to detect disrupted images. We use many image processing methods to obtain a set of augmented images. These augmented images are regarded as the protected input of deepfakes.
Correspondingly, there  are a lot of different representations  contained in the output.
With them, as shown in Figure~\ref{fig:MagDR} (c), we calculate the consistency score $\mathbf{S}_\mathrm{cons}$:
\begin{equation}\label{eq:6}
\begin{split}
    \mathbf{d}^k &=\mathbf{D}(\mathbf{G(\hat{x}, c)} , \mathbf{G}(\mathbf{f}_k(\mathbf{\hat{x}}),\mathbf{c})) \\
    \boldsymbol{\mu} &= \bm{\mathbf{E}}_{\mathbf{f}_k | \mathbf{F}}[\mathbf{d}^k] \\
    \boldsymbol{\sigma^2} &= \bm{\mathbf{E}}_{\mathbf{f}_k | \mathbf{F}}[(\mathbf{d}^k - \boldsymbol{\mu})^2] \\
    \mathbf{S}_\mathrm{cons} &= \boldsymbol{\sigma}
\end{split}
\end{equation}
Here, K is the number of image pre-processing methods.
If $\mathbf{S}_\mathrm{cons}$ is larger than the predefined threshold, we treat the input image as the adversarial example which may influence the whole image region, and vice versa.

\subsubsection{The Reconstructor}

The reconstructor aims to recover the correct output of deepfakes by reducing the effect of added perturbations. Formally, it is modelled as minimizing the distance between original output and reconstructed output:
\begin{equation}\label{eq:7}
    \min \mathbf{D(\mathbf{G(x, c), \mathbf{G(T(\hat{x}), c)}})}
\end{equation}
where $\mathbf{T(\cdot)}$ is an image transformation function. Ideally, $\mathbf{T(\cdot)}$ should be model-agnostic, sophisticated and  non-differentiable, making it harder for the adversary to circumvent the transformed model by back-propagating the distance metrics through it. 

As shown in Figure~\ref{fig:MagDR} (d), our proposed approach uses the image restoration technique to purify disrupted images. It has two components, which together form an effective pipeline that is difficult to bypass. First, we apply the conditional region mask to help us obtain specific facial patches. Second, we use a multi-stage module Rec-Net shown in Figure~\ref{fig:MagDR} (e), to enhance the image quality and simultaneously remove adversarial perturbations. Rec-Net is the core component of the Reconstructor, and its algorithmic description  is outlined in Algorithm.~\ref{algo:1}. 

\begin{algorithm}[h]

\footnotesize{\tcc{Pre-training Procedure}}
\footnotesize{\KwIn{Training set $\mathbf{X}$, Transformation methods $f_k^p$ for reconstruction in layer $\mathbf{B}_i$, $\mathbf{S}_\mathrm{final}$ refer to Eq.~\ref{eq:9}}}
\footnotesize{\KwOut{Layer $\mathbf{Q}_i$}}
\vspace{2mm}
\footnotesize{ $K$ = the number of method categories in $\mathbf{B}_i$.\\
$\mathbf{P}$ = the number of parameters for each category in $f_k$.\\
Evaluate each methods $f_k^p$ in training set $\mathbf{X}$ by $\mathbf{S}_\mathrm{final}$.\\
Find the optimal parameters $\mathbf{q}$ for each category methods $f_k$. \\ 
Insert $f_k^q$ into layer $\mathbf{Q}_i$.\\
$K$ = the number of methods in $\mathbf{Q}_i$}.\\ 

\vspace{2mm}
\tcc{Obtain the i-th transformation method for rec-net $\mathbf{R}$}
\footnotesize{\KwIn{Input image patch $x$, Layer $\mathbf{Q}_i$, Layer $\mathbf{B}_i$, $\mathbf{S}_\mathrm{final}$}}
\footnotesize{\KwOut{Processing methods sequence of rec-net $\mathbf{R}$}}
\vspace{2mm}

Evaluate each methods $f_k^q$ in layer $\mathbf{Q}_i$ for patch $x$.\\
Select the top-3 method categories $\mathbb{C}_j$ based on the $\mathbf{S}_\mathrm{final}$. \\
Use the top-3 categories as the index, obtain the subset layer $\mathbf{B'_i}$. \\
Select the top-1 scored method $f_{\mathbb{C}_j}^b$ in the subset as the optimal transformation method in layer $\mathbf{B_i}$. \\
Insert $f_{\mathbb{C}_j}^b$ into $\mathbf{R_i}$. \\

\caption{\footnotesize{ Rec-Net Training for a Patch}}
\label{algo:1}
\end{algorithm}

The final criteria score $\mathbf{S}_\mathrm{final}$  uses the distortion score $\mathbf{S}_\mathrm{dist}$ in Eq.~\ref{eq:8} and the consistency score $\mathbf{S}_\mathrm{cons}$ in Eq.~\ref{eq:6} to judge the abnormality of the output:
\begin{equation}\label{eq:9}
    \mathbf{S}_\mathrm{final} = \lambda \mathbf{S}_\mathrm{dist} + \mathbf{S}_\mathrm{cons}
\end{equation}
where $\lambda$ is the hyper-parameter to balance two different detectors.

\subsection{Advantages of Proposed Method}
Our proposed method offers a number of advantages.
First and most important, it is agnostic to  attack algorithms and  attacked models.
Second, as it leverages the vulnerability of perturbations, it thus can achieve strong detection efficiency of those perturbations with little altering.
Third, it takes strong adaptive defense ability with different attack degrees.
Fourth, unlike many recently proposed techniques, which degrade critical image information as part of their defense, our proposed method preserves image quality while simultaneously providing a strong defense.
Last, due to its modular nature, the proposed approach can be used as a universal module in existing deepfake models.

\begin{table*}[!tb]
\centering
  \caption{Comparison of detection performance across different attack methods and deepfakes. Note that all of these perturbations are generated under the defense-unaware situation.}
  \vspace{-3mm}
  \label{tab:1}
  \small
  \begin{tabular}{c c rrr rrr rrr}
    \toprule
  \multirow{3}{*}{Defense Models} &\multirow{3}{*}{Attack Methods} & \multicolumn{3}{c}{Face Replacement} & \multicolumn{3}{c}{Face Editing} & \multicolumn{3}{c}{Facial Reenactment} \\
    \cmidrule{3-11}
    & & Precision & Recall & F1  & Precision & Recall & F1 & Precision & Recall & F1 \\
    \midrule
    \centering\multirow{2}{*}{Lu \textit{et al.} \cite{lu2017safetynet}}
    & C\&W & 0.48 & 0.50 & 0.43 & 0.62 & 0.57 & 0.52 & 0.56 & 0.60 & 0.56 \\
    &PGD & 0.45 & 0.53 & 0.46 & 0.66 & 0.65 & 0.64 & 0.50 & 0.55 & 0.50 \\
    \hline
    \centering\multirow{2}{*}{Metzen \textit{et al.} \cite{metzen2017detecting}}
    & C\&W & 0.52 & 0.52 & 0.51 & 0.48 & 0.46 & 0.40 & 0.48 & 0.51 & 0.44 \\
    &PGD & 0.46 & 0.44 & 0.43& 0.50 & 0.55 & 0.49 & 0.48 & 0.47 & 0.40 \\
    \hline
    \centering\multirow{2}{*}{Meng \textit{et al.} \cite{meng2017magnet}}
    & C\&W & 0.51 & 0.51 & 0.50 & 0.53 & 0.53 & 0.52 & 0.49 & 0.50 & 0.46 \\
    &PGD & 0.56 & 0.56 & 0.56 & 0.90 & 0.89 & 0.89 & 0.56 & 0.55 & 0.55 \\
    \hline
    \centering\multirow{2}{*}{OTD}
    & C\&W & 0.28 & 0.53 & 0.36 & 0.20 & 0.33 & 0.25 & 0.60 & 0.57 & 0.46 \\
    &PGD & 0.27 & 0.52 & 0.35 & 0.52 & 0.51 & 0.50 & 0.45 & 0.49 & 0.36 \\
    \hline
    \centering\multirow{2}{*}{NNIF \textit{et al.} \cite{Cohen_2020_CVPR}}
    & C\&W & 0.82 & 0.66 & 0.73 & 0.83 & 0.85 & 0.84 & 0.85 & 0.54 & 0.66 \\
    &PGD & 0.78 & 0.62 & 0.69 & 0.83 & 0.79 & 0.81 & 0.80 & 0.52 & 0.63 \\
    \hline
    \centering\multirow{2}{*}{DD (ours)}
    & C\&W & 0.84 & 1.00 & 0.91 & 0.96 & 0.96 & 0.96 & 0.92 & 0.92 & 0.92 \\
    &PGD & 0.76 & 1.00 & 0.87 & 0.99 & 0.99 & 0.99 & 0.94 & 0.94 & 0.94 \\
    \hline
    \centering\multirow{2}{*}{CD (ours)}
    & C\&W & 0.87 & 0.99 & 0.92 & 0.96 & 0.96 & 0.96 & 0.76 & 0.86 & 0.81 \\
    &PGD & 0.95 & 0.94 & 0.94 & 0.98 & 0.99 & 0.99 & 0.97 & 0.78 & 0.87 \\
    \hline
    \centering\multirow{2}{*}{MagDR (CD+DD)}
    & C\&W & 0.96 & 1.00 & \textbf{0.98} & 0.96 & 0.96 & \textbf{0.96} & 0.92 & 0.92 & \textbf{0.92} \\
    &PGD & 0.95 & 1.00 & \textbf{0.97} & 1.00 & 1.00 & \textbf{1.00} & 0.97 & 0.94 & \textbf{0.95} \\
    \bottomrule
  \end{tabular}
\end{table*}

\section{Experiments}\label{sec:experiments}
\subsection{Datasets and Model Architecture}\label{sec:setting}
\textbf{Datasets.} We mainly use two datasets in our experiments: FaceForensics++~\cite{rossler2019faceforensics++} and CelebA~\cite{liu2015faceattributes}.
FaceForensics++ contains 1000 original video sequences. And all of them have been manipulated by Deepfakes, Face2Face, FaceSwap and NeuralTextures. The data is collected from 977 youtube videos and all videos contain a trackable mostly frontal face without occlusions.
CelebFaces Attributes Dataset (CelebA) contains 200K celebrity images that cover large pose variations and background clutter. CelebA has large diversities, large quantities, and rich annotations, including 10,177 number of identities, 202,599 number of face images, and 5 landmark locations, 40 binary attributes annotations per image. The size of each image is cropped to 128$\times$128.

\textbf{Model Architectures.} We use the CycleGAN, StarGAN, and GANimation image translation architectures to demonstrate our framework on different scenarios mentioned above. For CycleGAN, we use FaceForensics++ dataset to train a face to face model with 200 epochs. 
For StarGAN and GANimation, We use the open-source implementation~\footnote{\url{https://github.com/natanielruiz/disrupting-deepfakes}} refered in Nataniel \textit{et al.} \cite{ruiz2020disrupting} and fine-tuned on the CelebA dataset.

\subsection{Attack Settings and Evaluation Metircs}

We mainly use C\&W and PGD methods to craft adversarial examples. They are different attack methods in which one is gradient-based and another is optimization-based, which can prove our attack setting is comprehensive. 
We also use the different hyper-parameters, as well as the loss function to control the distortion in the different deepfake models. 
For CycleGAN attacking, we use the target attack settings in which the target label is the input image. And perform the untarget attack in StarGAN and GANimation models that could make more distortion of the output.
These objective functions can refer to Eq.~\ref{eq:1} and Eq.~\ref{eq:2}.

Specifically, all of those images are under the attack success situation and the magnitude of perturbation is constrained in the $\epsilon$ norm ball. 
And we adopt the \textit{recall} rate, the \textit{precision} rate and the \textit{F1} score to quantify the detection performance.
All experiments are run on the same set of images and against the same attacks for a fair comparison.

\subsection{Detection Performance}

We compare our proposed detector with a number of state-of-the-art adversarial examples detectors. These include training a model to distinguish the difference of adversarial examples and normal samples~\cite{lu2017safetynet, metzen2017detecting,Cohen_2020_CVPR}, calculating the reconstruction errors to detect adversarial examples~\cite{meng2017magnet}, and training a network to do binary classification on disrupted and clean outputs (OTD).

\subsubsection{Detecting Defense-unaware Attacks} 
We test the performance of three deepfakes under the defense-unaware attack, where the attackers generate the disrupted images in the models without any defense modules.
We calculate the precision, accuracy, and F1 score of detectors on disrupted images w.r.t. the perturbation $\epsilon$. 
Table.~\ref{tab:1} shows the performance of detectors under different attack methods and different deepfake scenarios.
From Table.~\ref{tab:1}, we can see that our method performs the best in all cases. Under the same attack settings, those methods proposed to detect adversarial examples in the classification tasks are at a low performance of detection.
And we also perform the \textbf{ablation study} to demonstrate the effectiveness of the consistency detector (CD) and distortion detector (DD). 
The results show consistency detector performs well in the face replacement.
Because the corruption in the situation is huge and influenced in the whole image, the distortion detector can not get a proper benchmark for comparison. 
And the distortion detector is good at detecting partial corruption. So it performs well in the facial reenactment.
Finally, MagDR combines the advantages of the two components to obtain superior detection performance.

\subsubsection{Detecting Defense-aware Attacks} 
\begin{figure}[!tb]
\centering\includegraphics[width=0.48\textwidth]{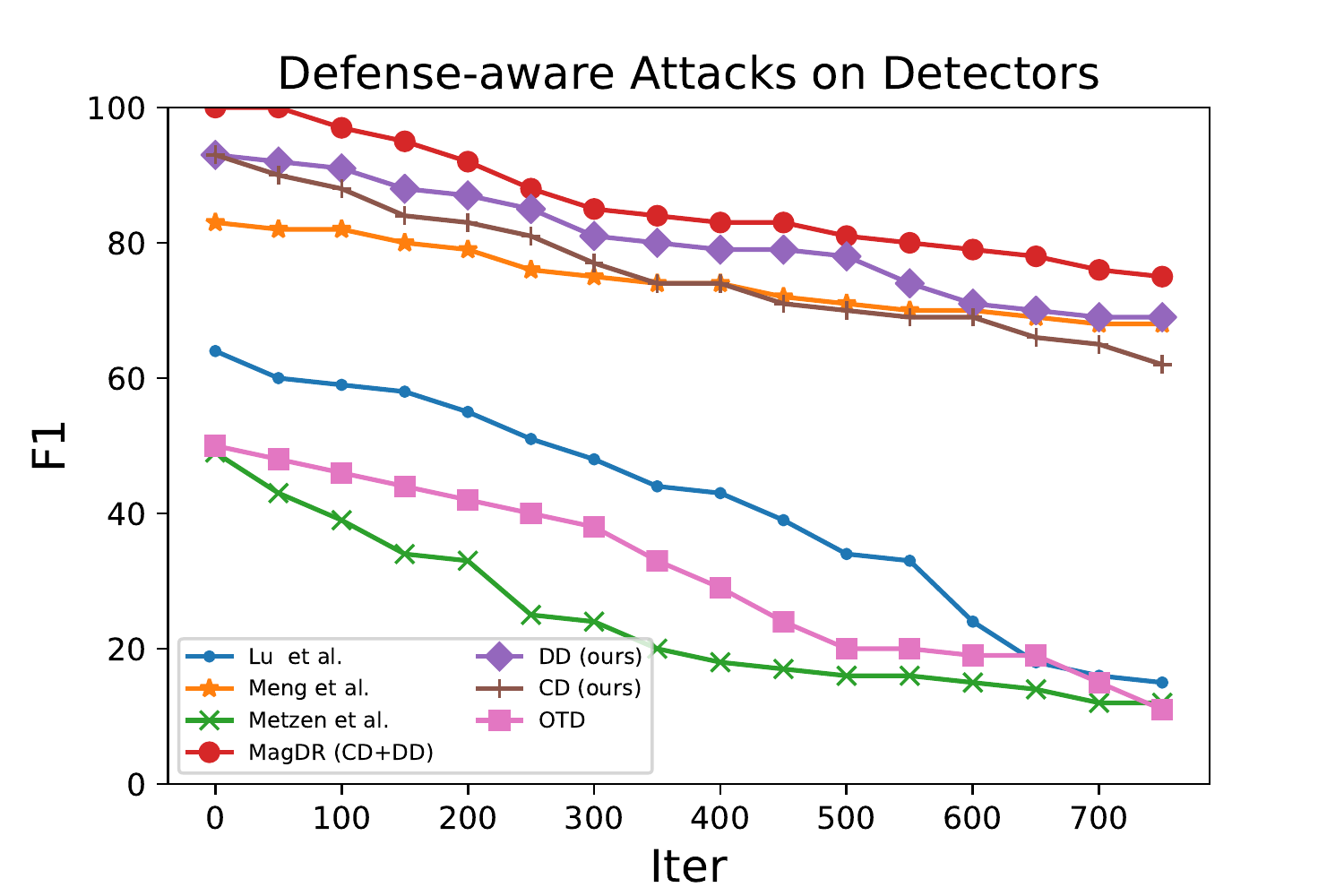}
\caption{Illustration the F1-score under defense-aware attacks. Iter denotes the iteration number of searching the adversarial perturbations.  }
\label{fig:adaptive_ack}
\vspace{-4mm}
\end{figure}

For a complete analysis, we investigate the detection performance under the defense-aware attack, which is also called adaptive attack~\cite{carlini2019evaluating}.
It is the most difficult defense scenarios because the adversary knows the
technique details of the detection methods. 
When launching an attack, the attacker can leverage the knowledge to fool the detector by generating specific perturbations.
The adaptive-attack methods can refer to Eq.~\ref{eq:3}, which ensures it can reduce the performance of detectors through more iterations. 
While a good detection method should increase the attack cost which means attackers should continuing to increase the iterations for the desired attack.
As Figure~\ref{fig:adaptive_ack} shows, with the iteration increase, the F1-score going down under the adaptive attack settings.
And the detectors which trained on some datasets shows their poor performance and high vulnerability.
While our method can greatly keep the stability of  highest detection performance under more aggressive attacks. 

\begin{table*}[!tb]
\centering
  \caption{Performance comparison with state-of-the-art reconstruction mechanisms. (I) is the difference between original input and reconstructed input. (O) is the difference between original outputs and reconstructed outputs.}
  \vspace{-3mm}
  \label{tab:2}
  \small
  \begin{tabular}{cccccccc|c}
    \toprule
\textbf{Metrics} &\textbf{Xie~\cite{xie2017mitigating}} & \textbf{Prakash~\cite{prakash2018deflecting}} & \textbf{Meng~\cite{meng2017magnet}}   & \textbf{Yang~\cite{yang2019me}}  &\textbf{Nataniel~\cite{ruiz2020disrupting}} & \textbf{Gintare~\cite{dziugaite2016study}} &\textbf{Mustafa~\cite{Mustafa_2020}} & \textbf{MagDR} \\
\hline
MSE (I) &25.59&25.56&27.66&21.17&24.59&17.69&15.53&\textbf{11.40}\\
SSIM (I) &0.72&0.72&0.76&0.81&0.75&0.82&0.88&\textbf{0.89}\\
PSNR (I) &32.90&32.91&30.12&32.44&33.25&36.11&38.60&\textbf{39.92}\\
Feature Similarity (I) &0.72&0.73&0.67&0.73&0.69&0.72&0.75&\textbf{0.77} \\
\hline
MSE (O) &257.19&242.34&34.98&27.47&43.64&35.25&30.60&\textbf{23.93}\\
SSIM (O) &0.09&0.08&0.68&0.75& 0.69& 0.82&0.86&\textbf{0.88}\\
PSNR (O) &12.86&13.37&28.08&30.17&28.26&30.12&31.35&\textbf{33.48}\\
Feature Similarity (O) &0.09&0.14&0.66&0.74&0.66&0.71&0.75&\textbf{0.76} \\
    \bottomrule
  \end{tabular}
\vspace{-2mm}
\end{table*}

\begin{figure*}[t]
	\begin{center}
		\centering
		\begin{minipage}[b]{0.19\linewidth}
			\centering
			\centerline{\includegraphics[width=0.98\linewidth]{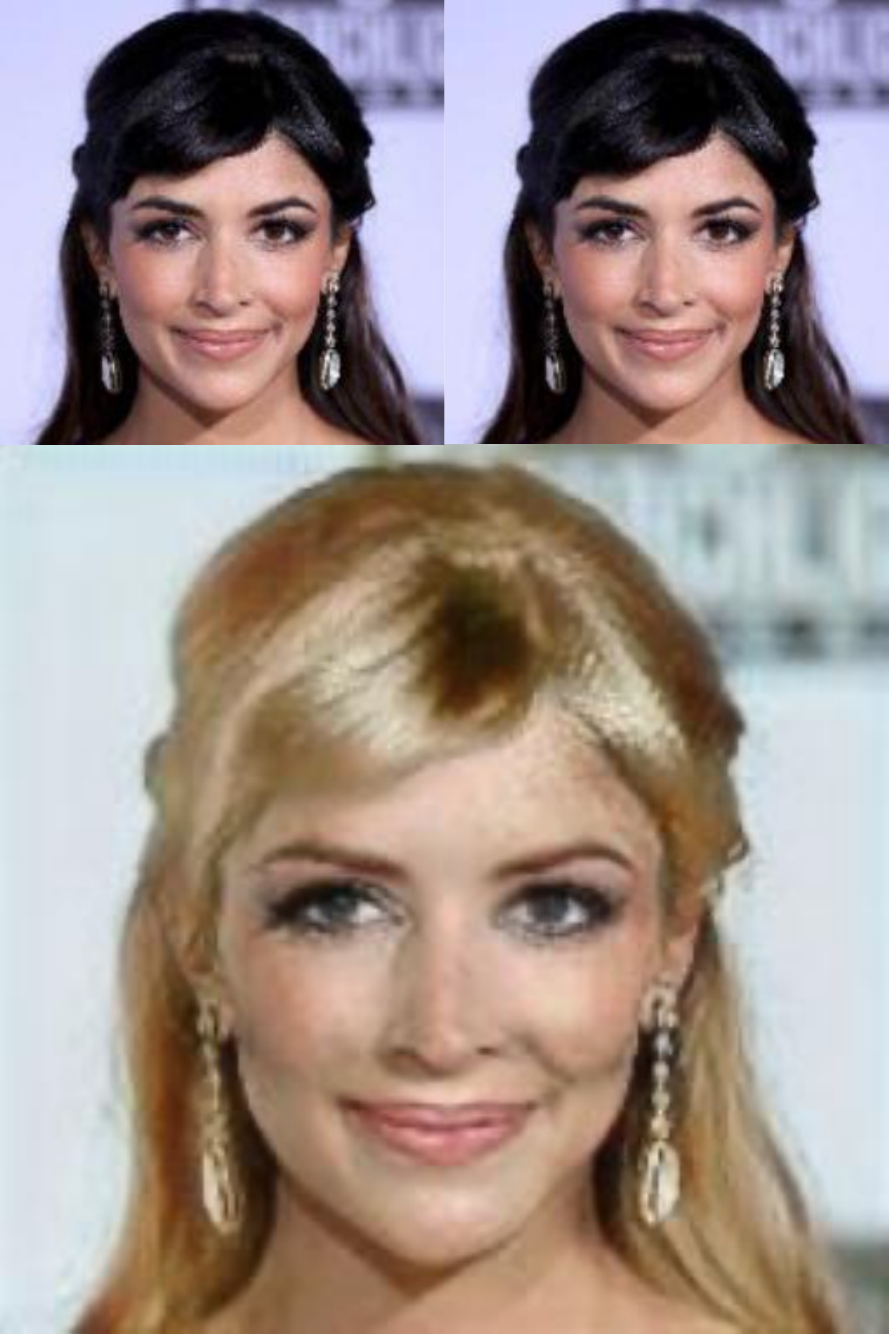}}
			\centerline{Ground-truth}
		\end{minipage}
		\begin{minipage}[b]{0.19\linewidth}
			\centering
			\centerline{\includegraphics[width=0.98\linewidth]{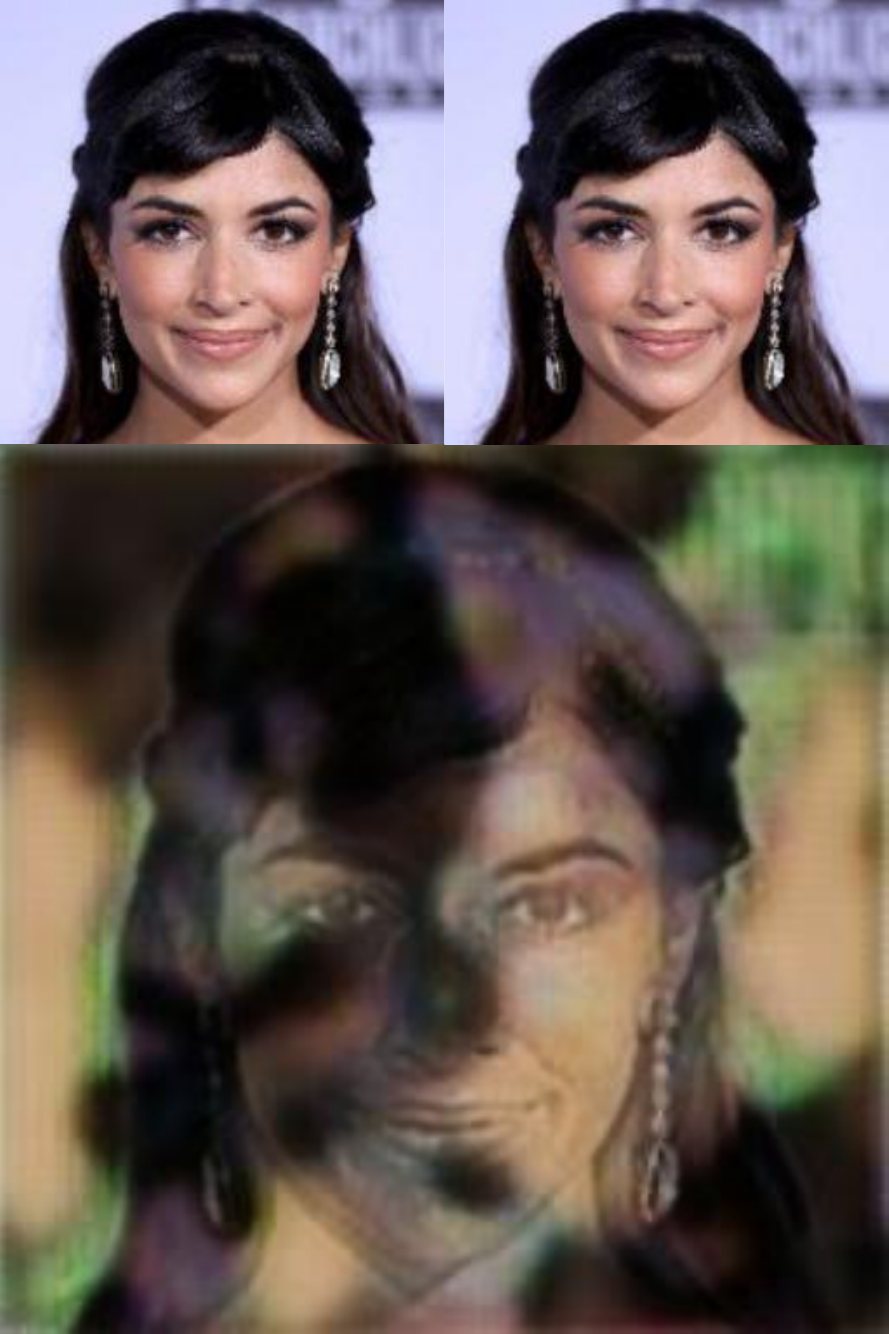}}
			\centerline{Without Defense}
		\end{minipage}
		\begin{minipage}[b]{0.19\linewidth}
			\centering
			\centerline{\includegraphics[width=0.98\linewidth]{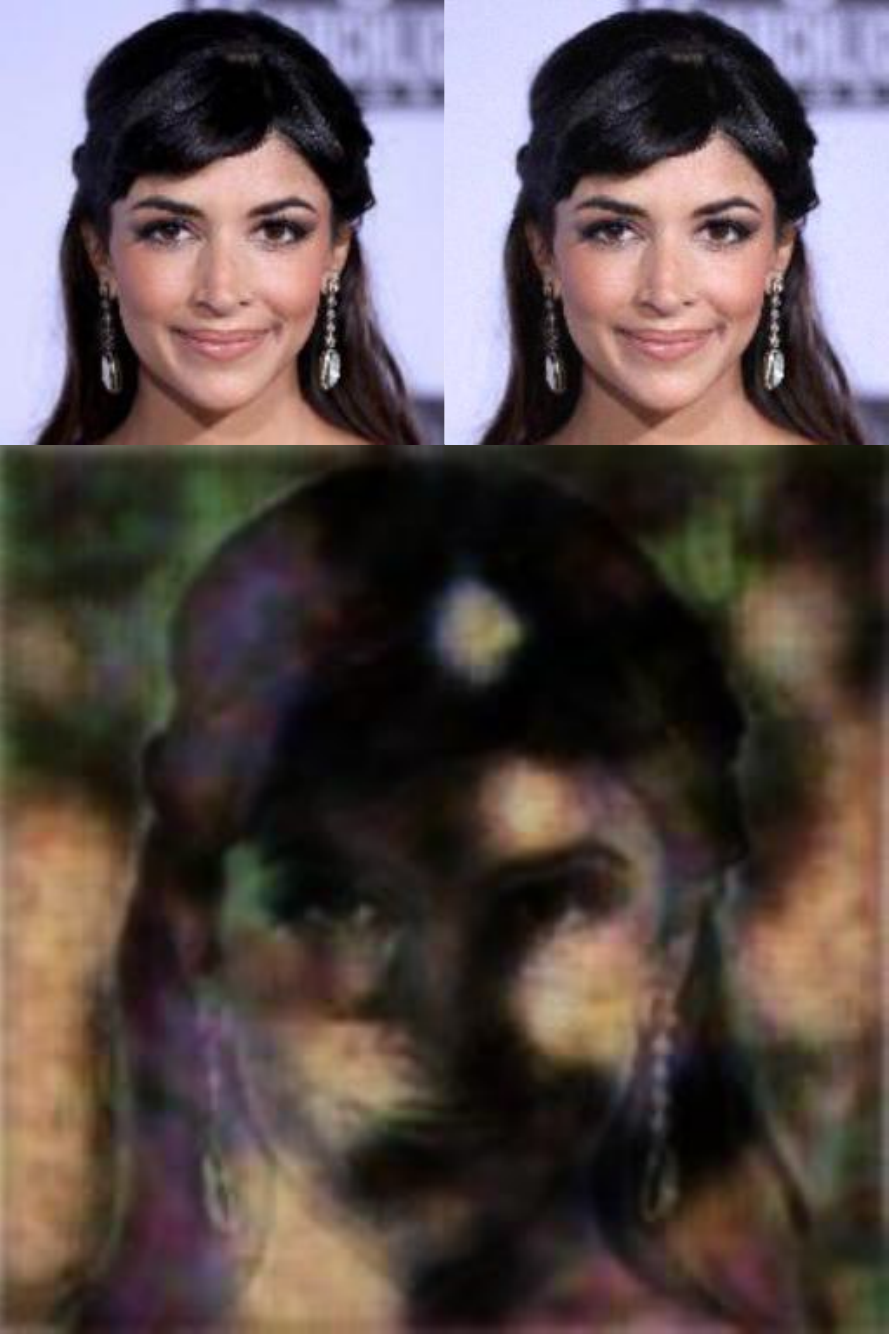}}
			\centerline{Xie \textit{et al.} \cite{xie2017mitigating}}
		\end{minipage}
		\begin{minipage}[b]{0.19\linewidth}
			\centering
			\centerline{\includegraphics[width=0.98\linewidth]{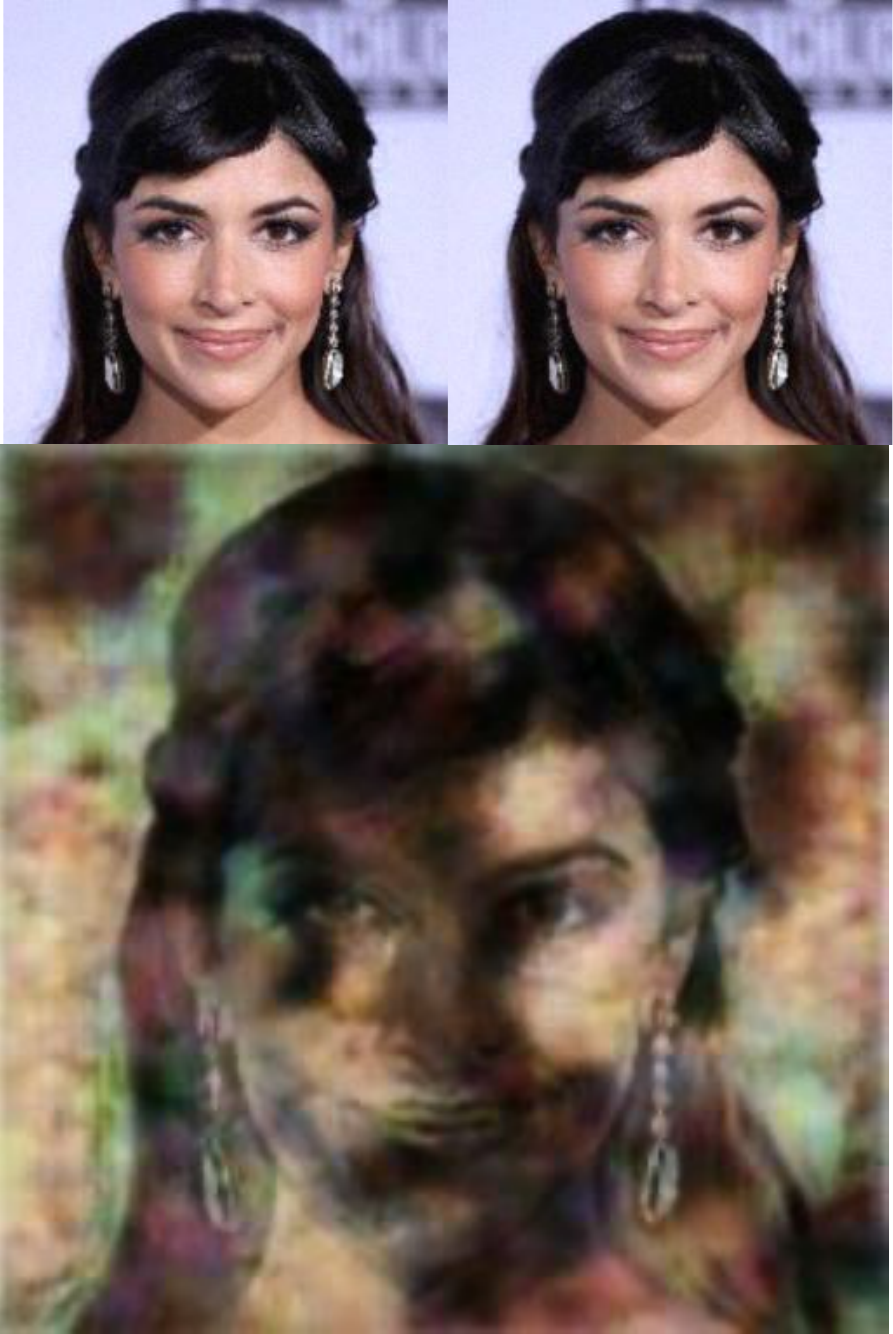}}
			\centerline{Prakash \textit{et al.} \cite{prakash2018deflecting}}
		\end{minipage}
		\begin{minipage}[b]{0.19\linewidth}
			\centering
			\centerline{\includegraphics[width=0.98\linewidth]{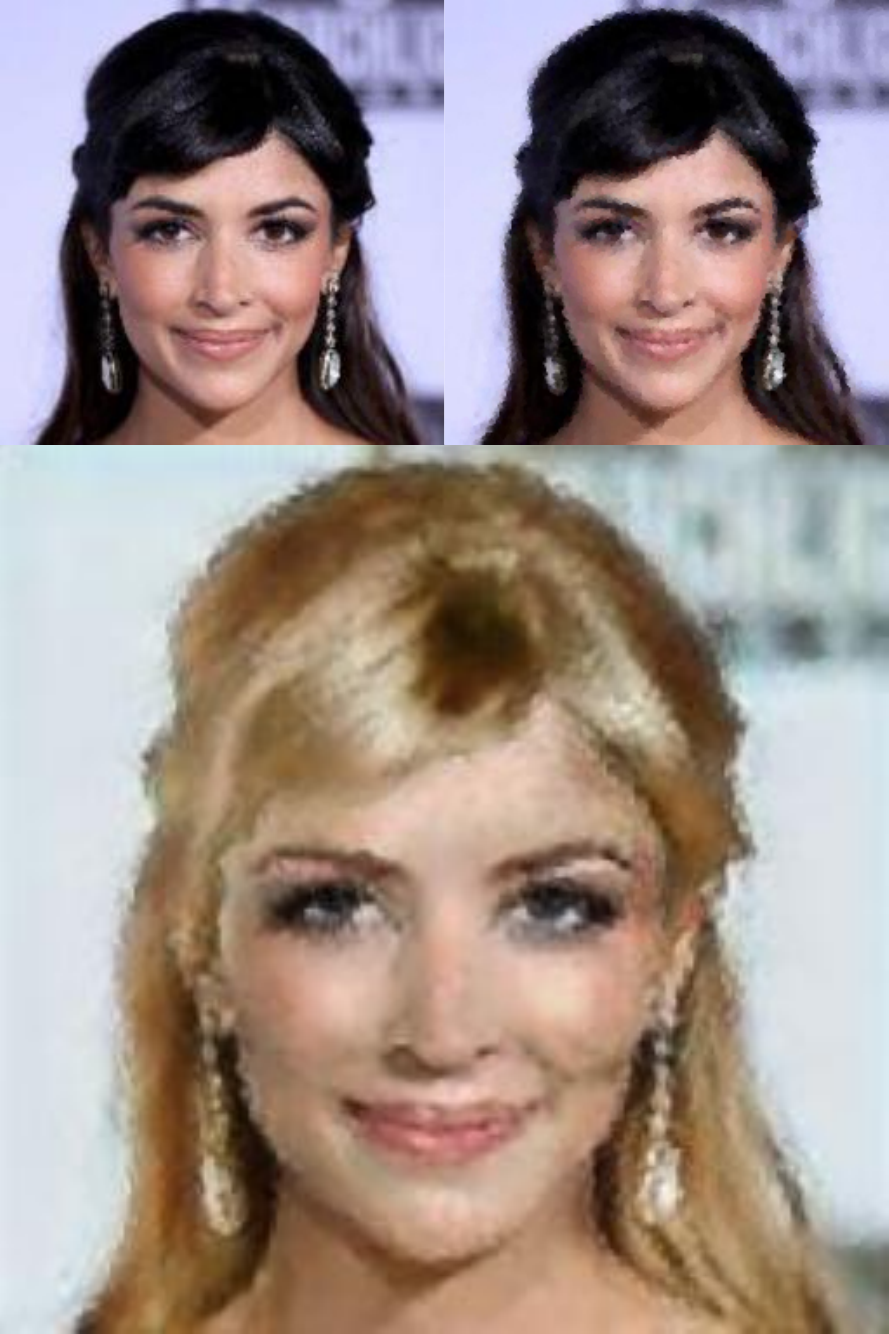}}
			\centerline{Meng \textit{et al.} \cite{meng2017magnet}}
		\end{minipage}
		\\ \medskip
		\begin{minipage}[b]{0.19\linewidth}
			\centering
			\centerline{\includegraphics[width=0.98\linewidth]{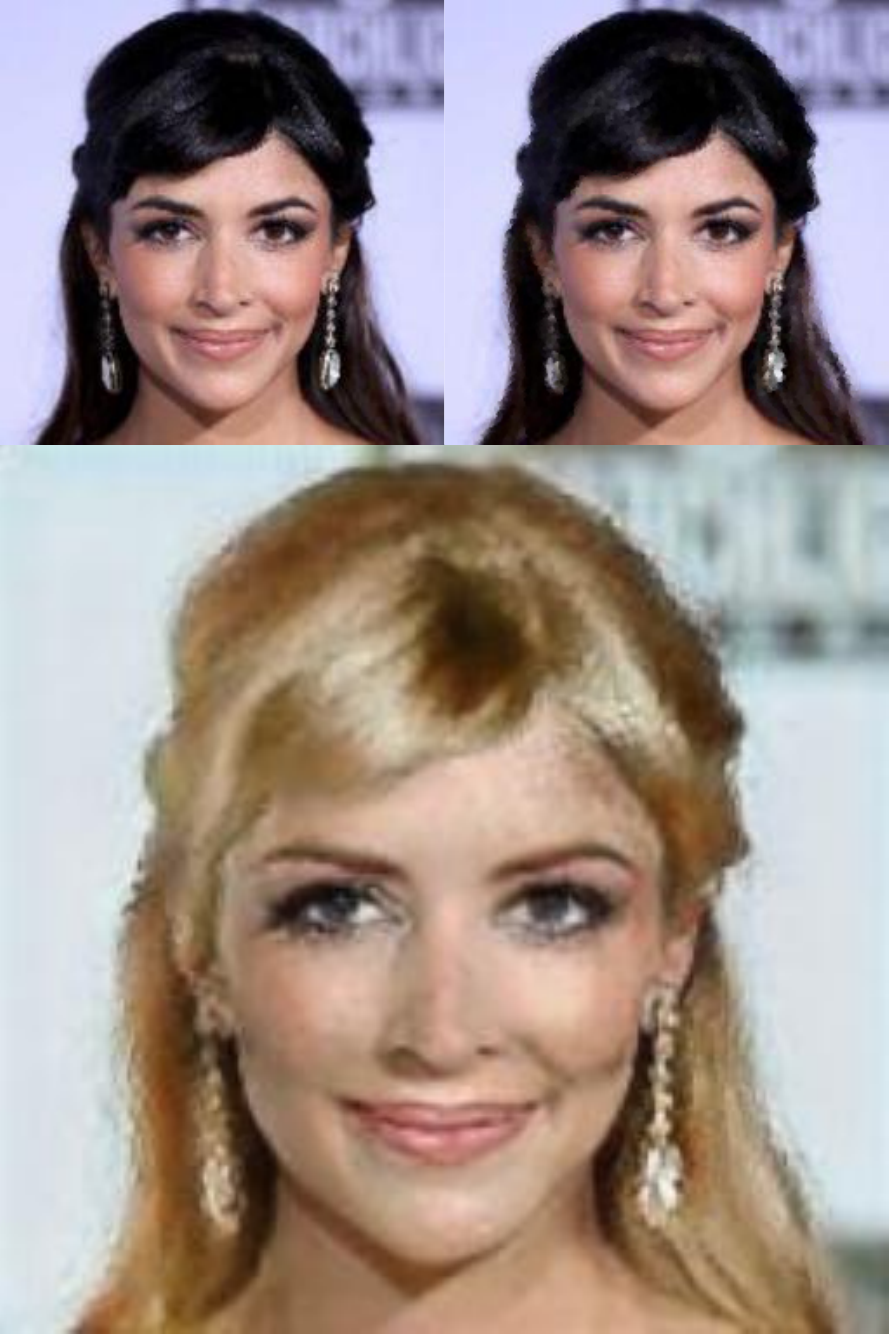}}
			\centerline{Yang \textit{et al.} \cite{yang2019me}}
		\end{minipage}
		\begin{minipage}[b]{0.19\linewidth}
			\centering
			\centerline{\includegraphics[width=0.98\linewidth]{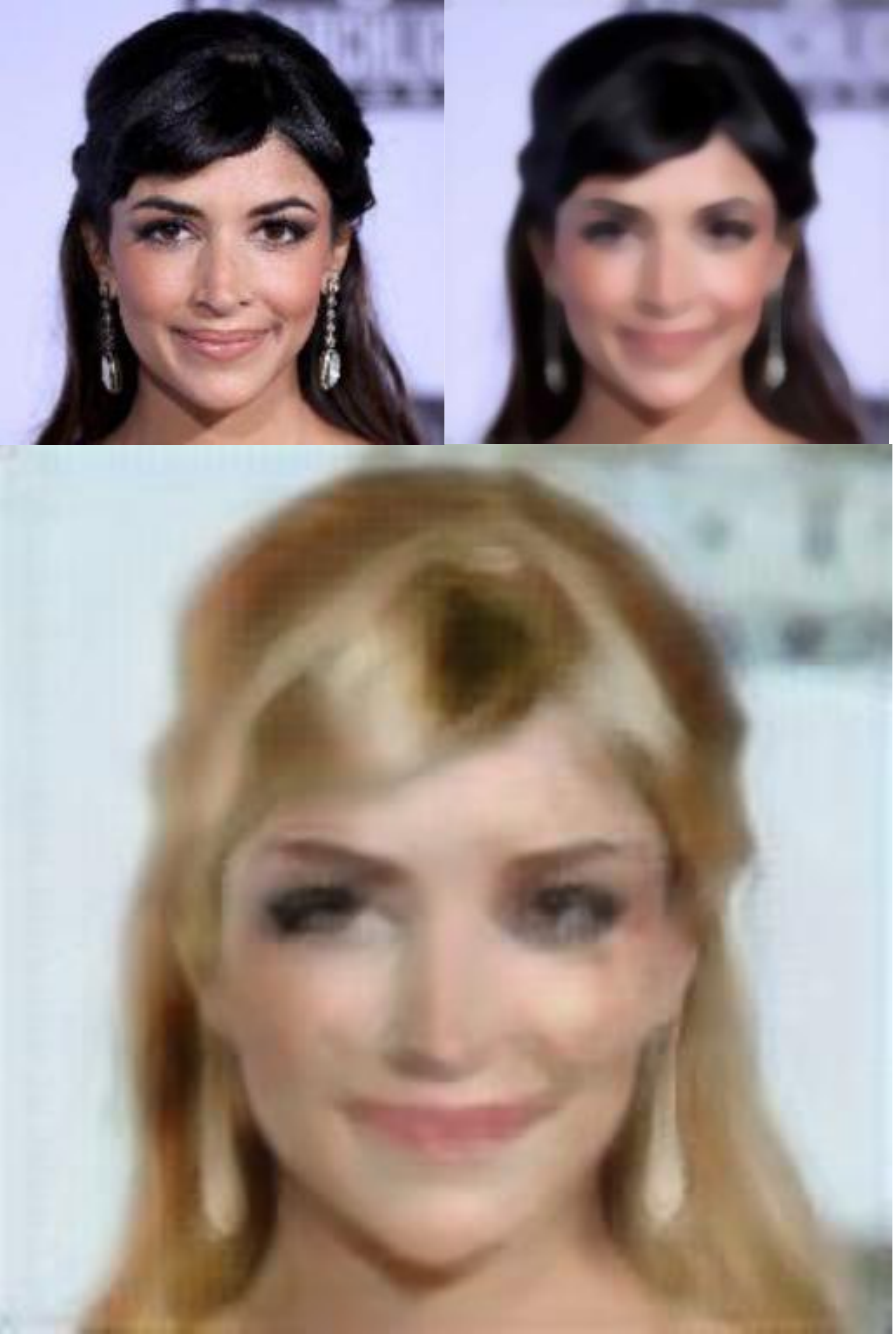}}
			\centerline{Nataniel \textit{et al.} \cite{ruiz2020disrupting}}
		\end{minipage}
		\begin{minipage}[b]{0.19\linewidth}
			\centering
			\centerline{\includegraphics[width=0.98\linewidth]{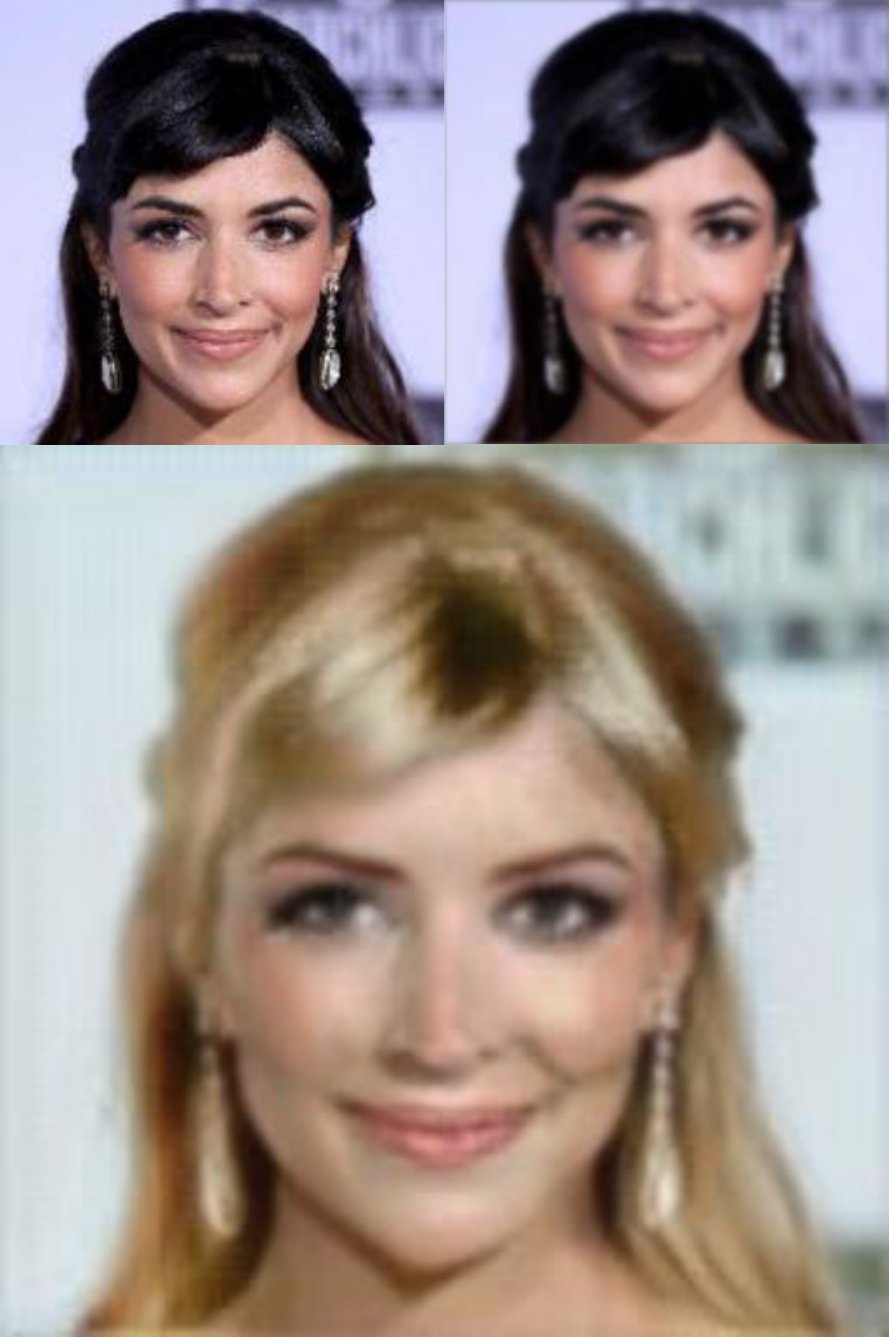}}
			\centerline{Gintare \textit{et al.} \cite{dziugaite2016study}}
		\end{minipage}
		\begin{minipage}[b]{0.19\linewidth}
			\centering
			\centerline{\includegraphics[width=0.98\linewidth]{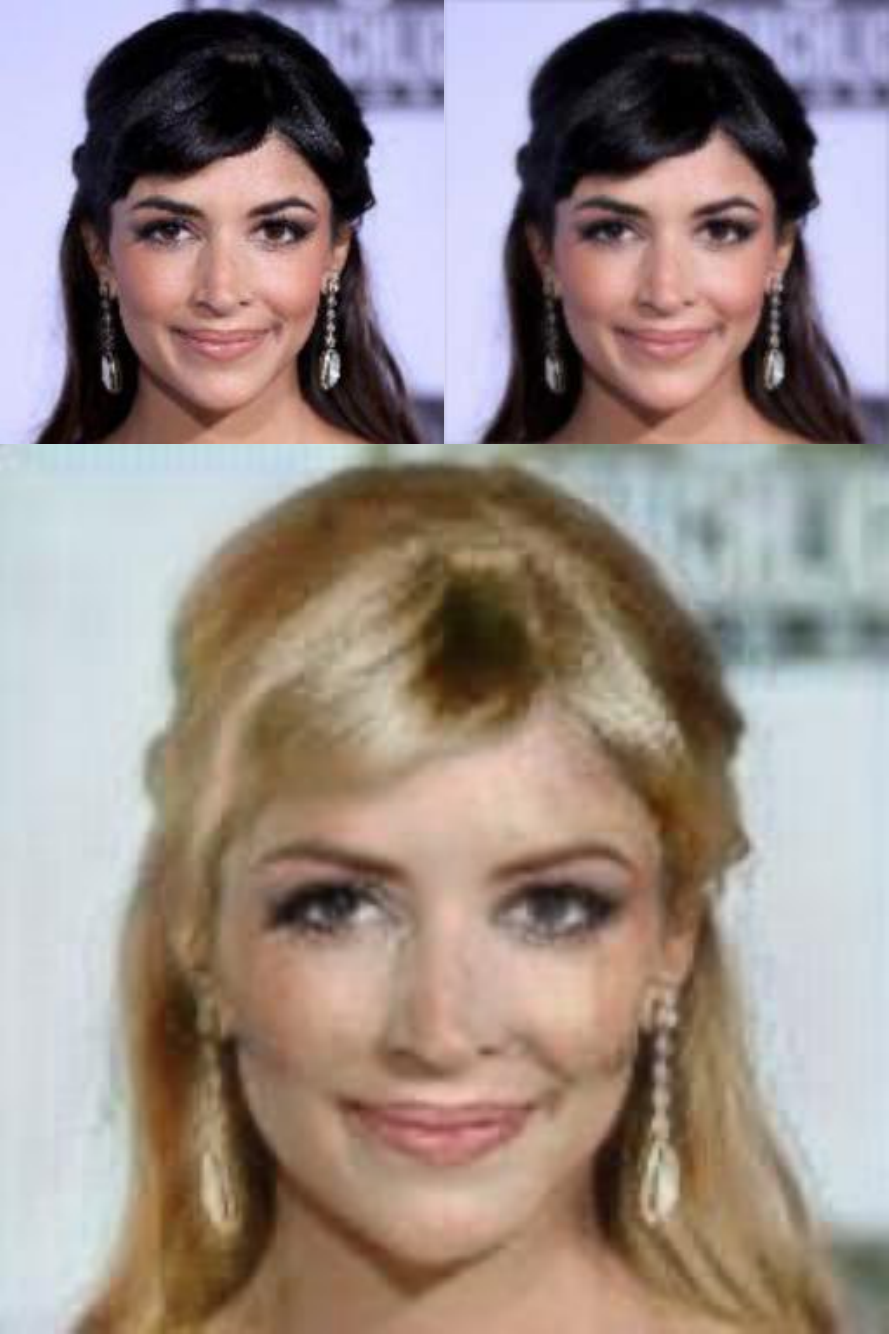}}
			\centerline{Mustafa \textit{et al.} \cite{Mustafa_2020}}
		\end{minipage}
		\begin{minipage}[b]{0.19\linewidth}
			\centering
			\centerline{\includegraphics[width=0.98\linewidth]{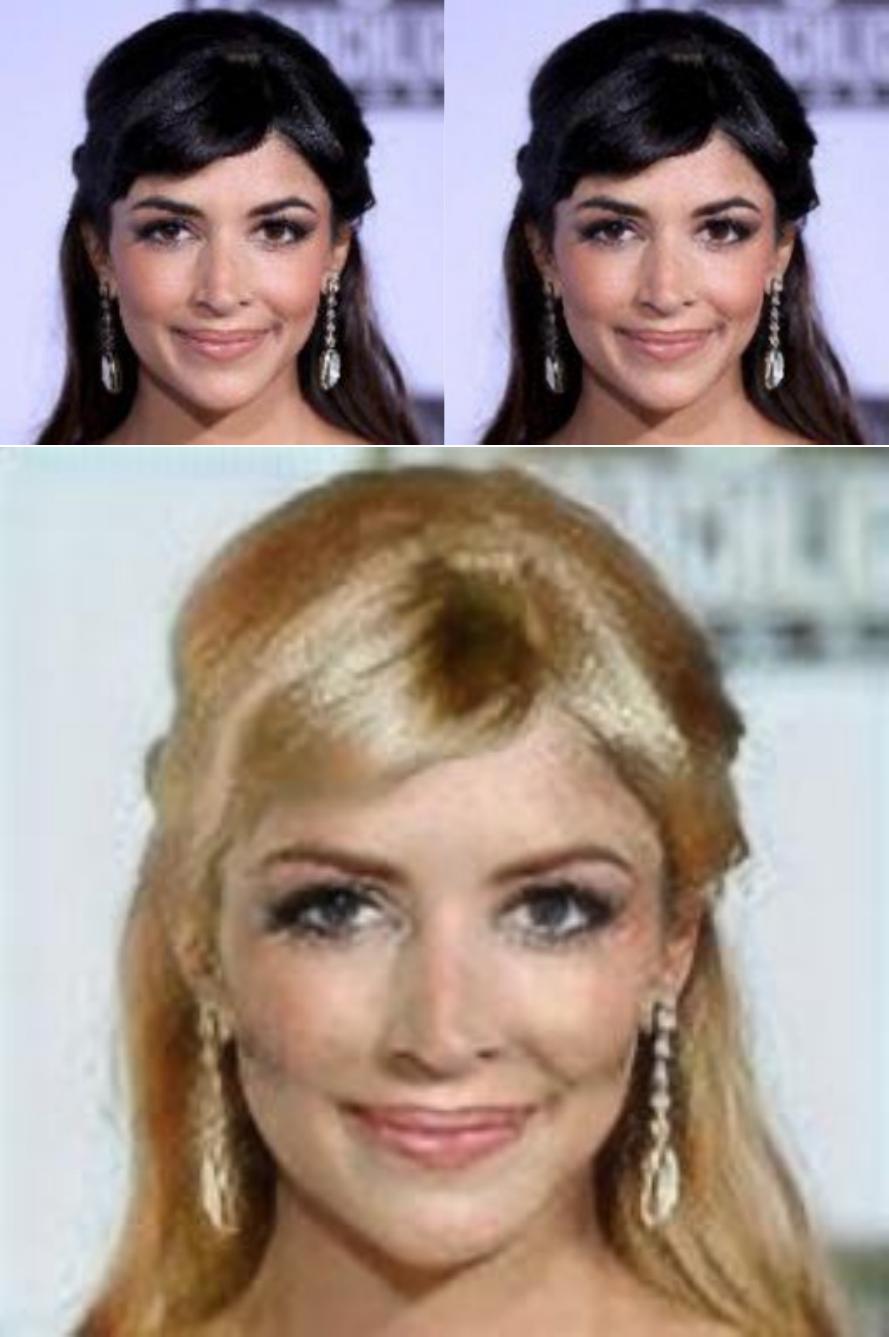}}
			\centerline{MagDR}
		\end{minipage}
	\end{center}
	\caption{Visual comparison of the deepfakes outputs of different reconstructed inputs by defense methods. In each case, (top-left) is the disrupted input, (top-right) is the reconstructed input, and (bottom) is the output obtained from the reconstructed input. The output images are enlarged two times for better visualization.}
	\label{fig:Rec_performance}
	\vspace{-4mm}
\end{figure*}

\subsection{Reconstruction Performance}

We compare our proposed reconstructor with number of recently introduced state-of-the-art image transformation based defense schemes in the literature. These include JPEG Compression~\cite{dziugaite2016study}, Adversarial training + Blur~\cite{ruiz2020disrupting}, Auto-encoder reformer~\cite{meng2017magnet}, Random Noise~\cite{xie2017mitigating}, Super-Resolution~\cite{Mustafa_2020},  Me-net~\cite{yang2019me}, Pixel Deflection(PD)~\cite{prakash2018deflecting}.

As Table.~\ref{tab:2} shows, we evaluate reconstructors under two dimensions: input-pair and output-pair. 
We expect the reconstructors should alter less in the input-pair while keeping high similarity in the output-pair.
The results show the randomized method~\cite{xie2017mitigating, prakash2018deflecting} does not have any reconstruction ability in deepfake tasks, even make things worse.
The Auto-encoder-based reconstructor~\cite{yang2019me, meng2017magnet} has a huge difference between the desired images.
And our method can perform superior in both two evaluated dimensions.

Figure~\ref{fig:Rec_performance} shows the effect of all of the compared defense methods on a disrupted image. The perturbations applied to samples are the same. And the reconstruction performance is quite equally with which is reflected in Table.~\ref{tab:2}.

\section{Conclusions}

This paper presents a two-step framework named MagDR (mask-guided detection and reconstruction) to defend deepfakes from adversarial attacks. The core idea is to compute a few unsupervised criteria that are sensitive to the adversarial perturbations on the output image. Then, an iterative process involving detection and reconstruction is performed, recovering the output to the desired form.

Beyond the promising results, our work delivers a message to the community that image-to-image translation algorithms seem easier to protect themselves from adversarial attacks because the attacks often generate meaningless patterns on the output image (rather than semantic predictions), making themselves easy to be detected. We expect the attacks to become stronger when they realize this weakness and generate more `natural' perturbations, and it may raise new challenges to defend such `smarter' attackers.

{\small

\bibliographystyle{ieee_fullname}
\bibliography{cvpr}
}

\end{document}